\documentclass[twoside]{article}

\usepackage[accepted]{aistats2025}
\usepackage[round]{natbib}

\usepackage{xcolor}
\usepackage{amsmath}
\usepackage{amsfonts}
\usepackage{algorithm,algpseudocode}
\usepackage{booktabs}
\usepackage{multirow}
\usepackage{multicol}
\usepackage{ulem}
\usepackage{microtype}
\usepackage{graphicx}
\usepackage{subfigure}
\usepackage{booktabs}
\graphicspath{ {figures/} }
\usepackage{hyperref}       
\usepackage{url}            
\usepackage{booktabs}       
\usepackage{amsfonts}       
\usepackage{nicefrac}       
\usepackage{xcolor}         
\usepackage{cleveref}
\usepackage[acronym]{glossaries}
\setkeys{glslink}{hyper=false}
\usepackage[nodisplayskipstretch]{setspace}
 \usepackage[inline]{enumitem}
 \usepackage{array}
\newtheorem{theorem}{Theorem}[section]

\newtheorem{conjecture}[theorem]{Conjecture}
\newtheorem{definition}[theorem]{Definition}

\newcommand{\ourmethod}{\textsc{SIRI}}

\begin{document}

%

%
\runningauthor{Bechler-Speicher, Eliasof, Schönlieb, Ran Gilad-Bachrach, Globerson}

\twocolumn[

\aistatstitle{On the Utilization of Unique Node Identifiers in Graph Neural Networks}

\aistatsauthor{ Maya Bechler-Speicher\textsuperscript{†1} \And Moshe Eliasof\textsuperscript{‡} \AND  Carola-Bibiane Schönlieb\textsuperscript{‡} \And Ran Gilad-Bachrach\textsuperscript{§¶} \And Amir Globerson\textsuperscript{†2}}
\vspace{2em}
\aistatsaddress{† Blavatnik School of Computer Science, Tel-Aviv University \\
‡ Department of Applied Mathematics and Theoretical Physics, University of Cambridge\\
§ Department of Bio-Medical Engineering , Tel-Aviv University\\
¶ Edmond J. Safra Center for Bioinformatics, Tel-Aviv University}  ]

\footnotetext[1]{Now also at Meta}
\footnotetext[2]{Now also at Google Research}
\footnotetext[3]{Email: mayab4@mail.tau.ac.il}
\begin{abstract}
Graph Neural Networks have inherent representational limitations due to their message-passing structure. Recent work has suggested that these limitations can be overcome by using 
unique node identifiers (UIDs). Here we argue that despite the advantages of UIDs, one of their disadvantages is that they lose the desirable property of permutation-equivariance. We thus propose to focus on UID models that are permutation-equivariant, and present theoretical arguments for their advantages. Motivated by this, we propose a method to regularize 
UID models towards permutation equivariance, via a contrastive loss. We empirically demonstrate that our approach improves generalization and extrapolation abilities while providing faster training convergence.
On the recent BREC expressiveness benchmark, our proposed method achieves state-of-the-art performance compared to other random-based approaches.

\end{abstract}

\section{Introduction}
\label{sec:intro}
Graph Neural Networks (GNNs), and in particular Message-Passing Graph Neural Networks (MPGNNs) \citep{morris2021weisfeiler} are limited in their expressive power, because they may represent distinct graphs identically \citep{garg2020generalizationrepresentationallimitsgraph}. This limitation can be addressed by assigning unique node identifiers (UIDs) to each node, thereby granting GNNs high expressiveness, and in the case of MPGNNs, they become Turing complete~\citep{loukas2020graphneuralnetworkslearn, cyclesgnns, abboud2021surprisingpowergraphneural}.
Given their substantial theoretical expressive power, UIDs are incorporated into various GNN frameworks~\citep{murphy2019relational, you2021identity,gpse_canturk, eliasof2024granola, papp2021dropgnnrandomdropoutsincrease}, with random node feature (RNF) augmentation \citep{ sato2021randomfeaturesstrengthengraph, abboud2021surprisingpowergraphneural} being one of the popular forms of UIDs.
However, the use of UIDs compromises permutation-invariance because, for two isomorphic graphs initialized with different random UIDs, the resulting GNN outputs may differ. Additionally, GNNs that use UIDs are prone to overfitting these IDs, which can negatively impact their generalization ability.

The introduction of random UIDs through RNF has also been explored using Random Node Initialization (RNI)\citep{sato2021randomfeaturesstrengthengraph, abboud2021surprisingpowergraphneural}  where the model resamples new RNFs in every forward pass of the network, and uses them as node features. Furthermore, \citet{abboud2021surprisingpowergraphneural}  have shown that GNNs with RNI are universal approximators of invariant and equivariant graph functions. 
Resampling RNFs is a heuristic intended to avoid the potential overfitting of the UIDs. Ideally, the resulting model should be invariant to the UIDs while offering improved expressiveness and downstream performance. Nonetheless, as shown in \cite{abboud2021surprisingpowergraphneural}, to learn UID invariance, the network requires substantially more epochs to converge compared with standard training of the baseline GNN before augmenting it with RNF.  
Additionally, it was shown that empirically, RNI does not consistently offer a performance boost on real-world and synthetic datasets \citep{bevilacqua2022equivariant, eliasof2023graph}.
 
In this paper, we alleviate the failure of GNNs to utilize UIDs.
Instead of relying on heuristics like RNI, we design a principled approach to learn invariance to UIDs, while preserving expressively.
We then theoretically analyze GNNs trained with UIDs and examine how to introduce UIDs-invariance to the model. We explore the idea of enforcing invariance at every layer but prove that such an approach does not enhance expressiveness compared to models without UIDs. Therefore, to improve expressiveness, it is necessary to include layers that are not invariant to UIDs. We also prove that $3$ layers, with only the last layer being invariant to UIDs, is enough to achieve expressiveness higher than 1-WL. 
Building on our theoretical insights, we show how UIDs utilization can be improved with explicit regularization using a contrastive loss.
We therefore call our approach \ourmethod~(\textbf{S}elf-supervised \textbf{I}nvariant \textbf{R}andom \textbf{I}nitialization), that is efficient and compatible with any GNN.
 We show that \ourmethod~significantly improves generalization, extrapolation, and accelerates training convergence. Notably, \ourmethod~outperform other expressive GNNs on the recent BREC expressiveness dataset~\cite {wang2024empiricalstudyrealizedgnn}.

\textbf{Main Contributions:} In this paper, we show how to improve the utilization of UIDs in GNNs. Our key contributions are as follows:
(1) We theoretically analyze the properties of GNNs with UIDs and derive concrete requirements to improve them.
(2) Based on our theoretical analysis, we show how to improve the utilization of UIDs using a contrastive loss, leading to \ourmethod --  an efficient method that is compatible with any GNN.
(3) We validate \ourmethod~on several expressiveness datasets, and show that it improves the ability of GNNs to utilize UIDs, reflected by its faster convergence and improved generalization and extrapolation downstream performance. 

\section{Related Work}
\label{sec:related}
In this section, we review the existing literature relevant to our study, focusing on the use of UIDs in GNNs, and the learning of invariacnes in neural networks.
\subsection{Unique Identifiers and Expressiveness in GNNs}
Node identifiers have emerged as a practical technique for enhancing the expressiveness of Graph Neural Networks (GNNs). In traditional GNN architectures, message-passing mechanisms aggregate and update node representations based on local neighborhood information, which inherently limits their ability to distinguish between nodes with structurally similar neighborhoods. This limitation makes GNNs only as powerful as the 1-WL (Weisfeiler-Lehman) test \citep{morris2021weisfeiler}, which fails to differentiate between certain graph structures.

To address this limitation, several works have introduced unique UIDs as additional node features during the message-passing process. By assigning unique  UIDs, GNNs can distinguish between nodes that are otherwise structurally identical, thus overcoming the limitations of the 1-WL test. The addition of UIDs enables the GNN to distinguish between isomorphic nodes that would otherwise be indistinguishable under traditional message-passing.
Some prominent examples of such techniques are presented in \citet{you2021identity} that injects UIDs to the node features. This approach demonstrated that adding such  UIDs could significantly enhance downstream performance on tasks like graph isomorphism and graph classification, where distinguishing between similar graph structures is crucial.

\subsection{Learning Invariances in Neural Networks}
Learning invariances in deep learning has been extensively explored in various domains, ranging from Convolutional Neural Networks (CNNs) to GNNs. 
In CNNs, translational invariance is inherently present due to the nature of convolutional operations. However, models like Spatial Transformer Networks (STN) enable CNNs to handle more complex geometric transformations, learning invariances to rotations, zoom, and more \citep{jaderberg2015spatial}. Additionally, the Augerino approach, which learns invariances by optimizing over a distribution of data augmentations, is an effective strategy for learning invariances in tasks like image classification \citep{benton2020learning}. TI-Pooling, introduced in transformation-invariant CNN architectures, further addresses the need for invariance across a variety of transformations in vision-based tasks \citep{laptev2016tipooling}.
In graph tasks, learning invariances has seen rapid development, particularly with graph neural networks (GNNs). Traditionally, GNNs are designed to be permutation invariant by construction, but recent research has also explored explicit mechanisms to learn invariances rather than relying solely on built-in properties. For instance, \citet{xia2023learning} introduced a mechanism that explicitly learns invariant representations by transferring node representations across clusters, ensuring generalization under structure shifts in the test data. This method is particularly effective in scenarios where training and test graph structures differ significantly, a problem known as out-of-distribution generalization. 
Another relevant approach is presented in \citet{jia2024graph}, who proposed a strategy to extract invariant patterns by mixing environment-related and invariant subgraphs. 
Differently, in \ourmethod, we consider learning UID invariance while promoting the use of its expressive power.

\section{Theoretical Analysis}
\label{sec:theory}

In this section, we establish theoretical results on GNNs with UIDs. Our goal is to examine how
to introduce UIDs-invariance to the model.
Due to space limitations, all proofs are deferred to the Appendix. 

\paragraph{Preliminaries} Throughout this paper, we denote a graph by $G = (V, E)$, and its corresponding node features by ${X} \in \mathbb{R}^{|V| \times d_0}$, where \( |V| \) is the number of nodes and \( d_0 \) is the input feature dimension.
When there are no node features, we assume that a constant feature $1$ is assigned to the nodes, as commonly done in practice~\citep{gin, morris2021weisfeiler}.
We consider GNNs with \( L \) layers. The final layer is a classification/regression layer, depending on the task at hand, and is denoted by $g$.
In favor of clarity, we assume the UIDs are randomly generated by the model and augmented as features. We will refer to such GNNs as \textit{GNN-R}, and to GNNs that do not generate UIDs as \textit{regular GNNs}.
We use the following definition for UIDs-invariance:
\begin{definition}
    A function $f$ is \textit{UIDs-invariant} if 
    $\mbox{Var}[f(X;UID)] = 0$ where UID is sampled from some distribution $P$.
\end{definition}
Here $;$ denotes the concatenation operation, and UID is a matrix of fixed-size random feature vectors.

First, we note that a GNN-R that is not UIDs-invariant always has a non-zero probability of failing with some error at test time,
Therefore, we prefer solutions that are UIDs-invariant. 

Another source of complication is that a function can be UIDs-invariant with respect to some graphs, but non-UIDs-invariant with respect to others. This is stated in the following theorem:
\begin{theorem}\label{thm:invarinace_can_be_bad}
    A function $f$ can be UIDs-invariant with respect to a set of graphs $S$ and non-UIDs-invariant with respect to another set of graphs $S'$.
\end{theorem}
To prove Theorem~\ref{thm:invarinace_can_be_bad} we construct a UIDs-invariant function that is graph-dependent. 
A direct implication of Theorem~\ref{thm:invarinace_can_be_bad} is that it is possible for a GNN-R to be UIDs-invariant on training data, but non-invariant on test data, leading to high test error.

To design a method that effectively leverages UIDs, we must understand the limitations of a GNN-R that is UIDs-invariant. One question is how to introduce invariance to the model. 
The next theorem shows that enforcing invariance to UIDs within every GNN layer does not enhance the model's expressiveness at all, at least for Message-Passing GNNs.

\begin{theorem}\label{thm:mpnnwithidsepxressivity}
Let $G$ be the set of regular Message-Passing GNN models, and $G'$ the set of Message-Passing GNN-R models that are UIDs-invariant in every layer, with the same parameterization as $G$. Then the set of functions realized by $G$ and $G'$ are the same.
\end{theorem}

To prove Theorem~\ref{thm:mpnnwithidsepxressivity} we show by induction that any function in $G'$ can be computed by some function in $G$ (the reverse containment is clear).

Theorem~\ref{thm:mpnnwithidsepxressivity} implies that in order to design a network that is both UIDs-invariant and expressive, we must allow it to be non-UIDs-invariant in at least one hidden layer.
The next theorem shows that three layers and only enforcing invariance in the last layer of Message-Passing GNNs, already provide the network with expressive power higher than 1-WL.
\begin{theorem}\label{thm:only_last_layer}
There exist functions that no Message-Passing GNN can realize and can be realized by a UIDs-invariant GNN-R with $3$ layers where only the last layer is UIDs-invariant.
\end{theorem}

To prove Theorem~\ref{thm:only_last_layer} we construct a UID-invariant solution for the task of detecting a triangle.

Finally, we state what is the expressive power limitation of a GNN-R that is UIDs-invariant. 

\begin{theorem}\label{thm:isomorphism}
A GNN-R that is UIDs-invariant and runs in polynomial time, is not a universal approximator of equivariant graph functions, unless graph isomorphism is NP-complete.
\end{theorem}

Theorem \ref{thm:isomorphism} follows from the fact that a GNN-R that is UIDs-invariant, can solve graph isomorphism, which is believed to be NP-intermediate \citep{chen2023equivalencegraphisomorphismtesting}.

In the next section, we build upon the theoretical results discussed in this section, to design an approach that enhances the utilization of UIDs in GNNs by explicitly enforcing UIDs-invariance to the model.

\section{Learning Invariance to UIDs}
\label{sec:method}

In this section, we build upon our theoretical results shown in Section~\ref{sec:theory}, to enhance the ability of GNNs to utilize UIDs.
Specifically, Theorem~\ref{thm:mpnnwithidsepxressivity} indicates that enforcing invariance at every layer of the network is not only unnecessary but also does not contribute to the model’s expressive power. Furthermore, Theorem~\ref{thm:only_last_layer} reveals that enforcing invariance solely at the final layer is sufficient to achieve UIDs-invariance while simultaneously improving the network's expressiveness.
Based on these insights, we propose explicitly enforcing invariance at the network's last layer using a residual contrastive loss. By avoiding unnecessary invariance constraints in the intermediate layers, our approach maintains and enhances the model’s capacity to differentiate and leverage UIDs effectively. We refer to our proposed method as \ourmethod.

We consider a GNN layer of the following form:
\begin{equation*}
    \label{eq:general_gnn}
{H}^{(l+1)} = \textsc{GNN}_{\theta^{(l)}}({H}^{(l)}; G),
\end{equation*}
where $\textsc{GNN}_{\theta^{(l)}}$ is the $l$-th layer, and it depends on the input graph structure $G$ and the latest node features ${H}^{(l-1)}$. The layer is associated with a set of learnable weights \( \theta^{(l)} \) at each layer \( l \in \{0, 1, \dots, L-1\} \). The initial node features $H^{(0)} \in \mathbb{R}^{n \times (d_0+r)}$ are composed of the concatenation of input node features $X$ and random UIDs $R \in \mathbb{R}^{n \times r}$ sampled at each forward pass of the network such that the hidden dimension of the network layers is $d=d_0+r$. That is, $H^{(0)} = 
X;R$, where $;$ denotes channel-wise concatenation.

\begin{figure*}[t]
    \centering

      \subfigure[Train]{\label{figure:siri_vs_rni_train_ratio.} \includegraphics[width=0.45\textwidth]{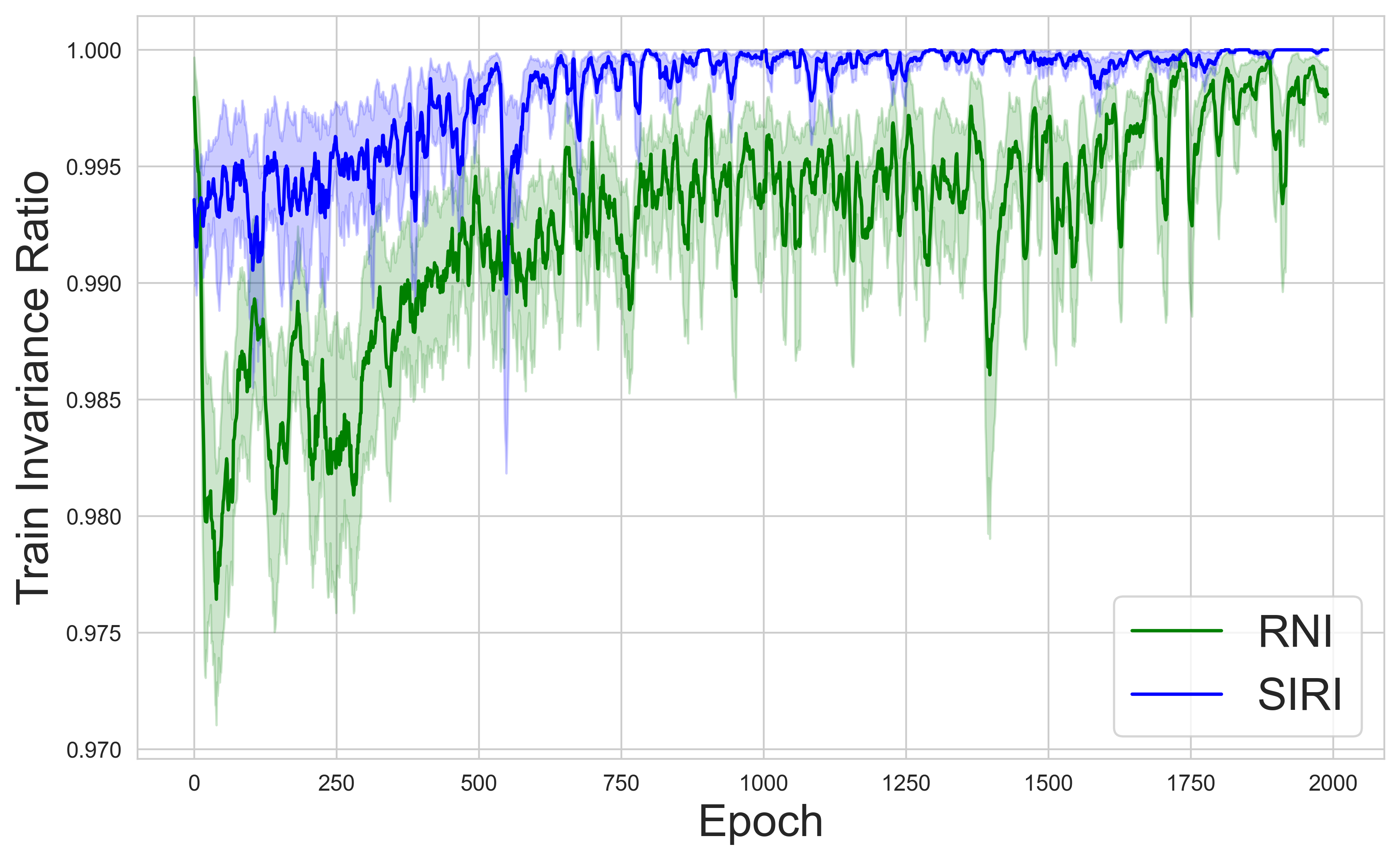}}\quad
      \subfigure[Test]{\label{figure:siri_vs_rni_test_ratio}\includegraphics[width=0.45\textwidth]{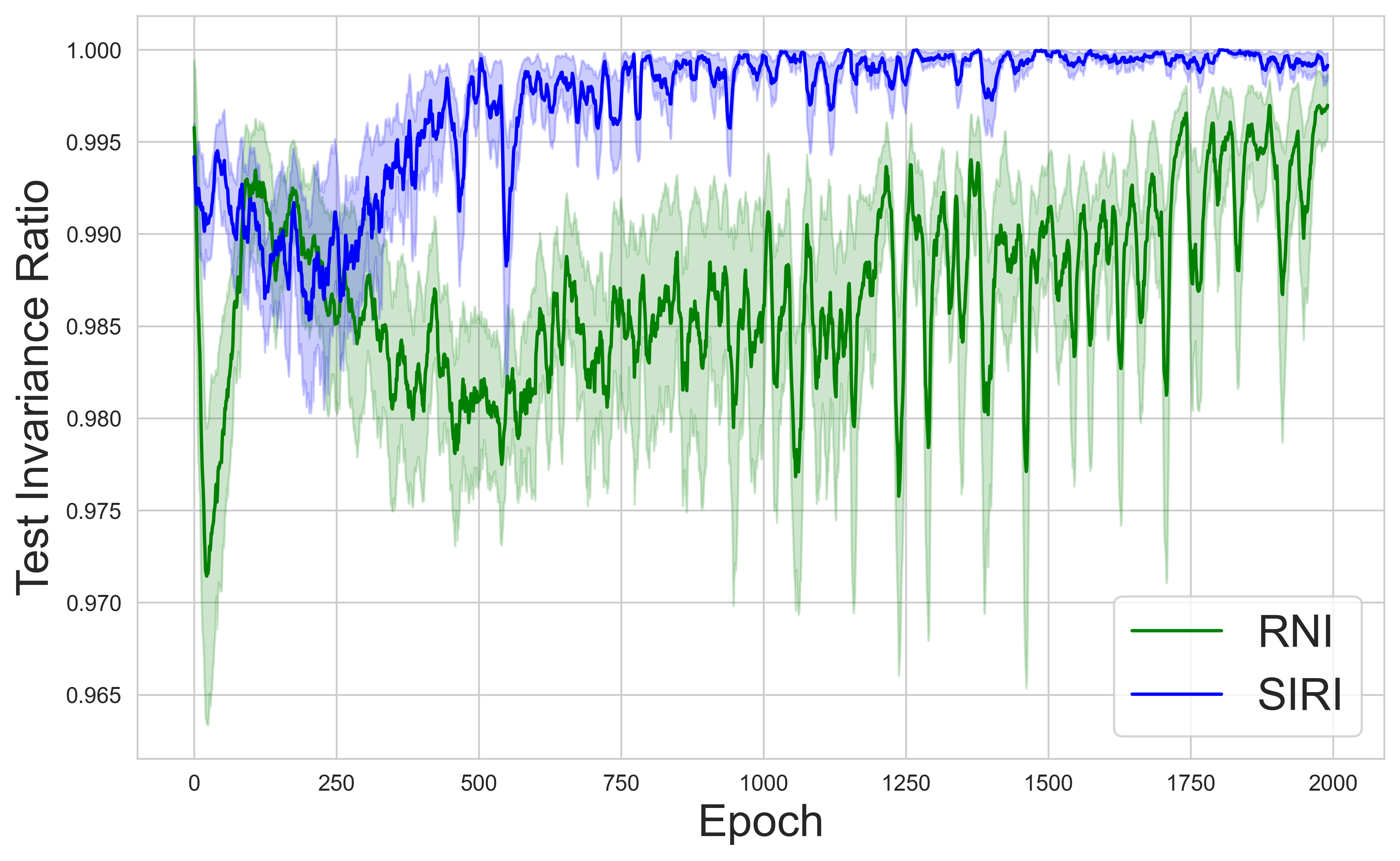}}

    \caption{The UID invariance ratio curves obtained with SIRI compared to RNI, with respect to the train set (a) and test set (b).}
    \label{Figure:siri_vs_rni}
\end{figure*}
Overall, \ourmethod~ consists of two mechanisms: (1) a contrastive loss that promotes the invariance to UIDs, and (2) a furthest RNF optimization approach, that seeks to optimize pairs of RNFs that yield different final node embeddings, as described below.

\paragraph{Task Loss ($ \mathcal{L}_{\text{task}}$).}
Given a downstream task, such as graph classification or regression, we compute the task loss $ \mathcal{L}_{\text{task}}$ based on the output of the final GNN layer ${H}^{(L)}$. Specifically, we use a standard linear classifier $g: \mathbb{R}^{d} \rightarrow \mathbb{R}^{d_{\rm{out}}}$, where ${d_{\rm{out}}}$ is the desired dimension of the output (e.g., number of classes), to obtain the final prediction $\hat{y} = g({H}^{(L)})$. Following that, we compute the considered loss (e.g., mean-squared error or cross-entropy). Formally, the task loss reads:
\begin{equation*}
    \label{eq:task_loss}
\mathcal{L}_{\text{task}} = \text{Loss}( \hat{y}, y), 
\end{equation*}
where \( y \) is the ground truth label or value.

\textbf{Contrastive Loss ($\mathcal{L}_{\text{contrastive}}$).}
To compute the contrastive loss, we generate two versions of the input graph and its features, \( G_1 = (V, E, {X}, {R}_1) \) and \( G_2 = (V, E, {X}, {R}_2) \), where \( {R}_1 \) and \( {R}_2 \) are two different RNFs, sampled on-the-fly. We propagate both inputs through the network up to the last GNN layer $L-1$ (prior to the final classifier), and compute their embeddings \( {H}_1^{(L)} \) and \( {H}_2^{(L)} \). To promote UIDs-invariance, we consider a contrastive loss that aims to reduce the difference between these two \textit{numerically} different, yet \textit{semantically} equivalent representations of the graph. Namely, we compute the mean squared error (MSE) between the two embeddings, as follows:

\begin{equation*}
\label{eq:residual_loss}\mathcal{L}_{\text{contrastive}} = \text{MSE}({H}_1^{(L)},  {H}_2^{(L)} ).
\end{equation*}

The overall loss to optimize is:
\begin{equation*}
    \mathcal{L} = \mathcal{L}_{\text{task}} +  \mathcal{L}_{\text{contrastive}}.
\end{equation*}

 Depending on running time limitations, it is also possible to further optimize the selection of ${R}_2$. To that end, we sample $k$ vectors of ${R}_2$ and apply the gradient update with the one that maximizes $\mathcal{L}_{\text{contrastive}}$.

\section{Experimental Evaluation}
\label{sec:experiments}

In this section, we evaluate \ourmethod~on several synthetic benchmarks. Note that, following \citet{abboud2021surprisingpowergraphneural} and \citet{wang2024empiricalstudyrealizedgnn}, we choose to work with these benchmarks because they are designed to assess the expressive power of GNNs. Therefore, achieving satisfactory performance on these datasets sheds light on the ability of \ourmethod~to improve the expressiveness of 1-WL GNNs by effectively utilizing UIDs. Each of the following subsections focuses on a different research question, from the improved generalization and extrapolation capabilities of \ourmethod, to its improved expressiveness.

\paragraph{Baselines} To establish a comprehensive and directly related set of baselines to assess the performance of our \ourmethod, we focus on random-based methods. Specifically, we consider (1) DropGNN \citep{papp2021dropgnnrandomdropoutsincrease}, which randomly drops edges in input graphs at each forward pass; (2) OSAN \citep{qian2022ordered}, a randomized subgraph GNN method; and (3) RNI \citep{abboud2021surprisingpowergraphneural}, which augments node features with RNFs resampled at every forward pass of the network.

\paragraph{Experimental Settings} To ensure a fair evaluation, in all experiments considering RNI and our \ourmethod, we use  GraphConv \citep{morris2021weisfeiler} as the GNN backbone, because it is the most expressive MPGNN among the 1-WL expressive MPGNNs \citep{morris2023weisfeiler}. Furthermore, for the comparison of \ourmethod~with RNI, we perform two random samplings of RNF in each step in RNI. This is because in \ourmethod, in every step, there are two randomly sampled RNF - one as in RNF in the forward pass, and another one from the residual loss. 
We provide additional details on our experimental settings, including information on the experimental choices and hyperparameter tuning in \Cref{app:experimental_settings}.

\subsection{Improving Generalization and Extrapolation}

In this subsection, we examine the generalization and extrapolation performance of SIRI with respect to RNI, as well as their invariance properties. We use a synthetic dataset where we can guarantee the network can solve.

\paragraph{Dataset}
The isInTriangle task is binary node classification: determining whether a given node is part of a triangle.
The dataset consists of 100 graphs with 100 nodes each, generated using the preferential attachment (BA) model \citep{badist}, in which graphs are constructed by incrementally adding new nodes with $m$ edges, and connecting them to existing nodes with a probability proportional to the degrees of those nodes. We adopt an inductive setting, where the graphs used during testing are not included in the training set. 
We used $m=2$ for the training graphs and evaluate two test sets: an interpolation setting where the graphs are drawn from the BA distribution with $m=2$, and an extrapolation setting where the graphs are drawn from the BA distribution with with $m=3$.
The train set and test set consists of $500$ nodes each.

\paragraph{Setup}
We trained a 6-layer GraphConv MPGNN \citep{morris2021weisfeiler} for 2000 epochs to ensure convergence and 64 hidden dimensions.
We conducted two experiments:
\newline
\newline
(1) Baseline model: All nodes were assigned a constant feature value of 1.
\newline
 (2) Model with RNI: The network uses RNI with 64 RNF dimensions.
\newline
(3) Model with \ourmethod: The network uses SIRI with 64 RNF dimensions.

 We use GraphConv as, according to \citet{cyclesgnns}, in the baseline scenario, MPGNNs are incapable of detecting cycles. The baseline evaluation allows us to assess the effectiveness of incorporating UIDs through RNI.
 Our objective is to examine whether the network utilizes the UIDs and, if so, whether it learns a solution that is invariant to their values.

To evaluate the network’s invariance to UIDs in both the training and test sets, we employed the following procedure. For each node in a given set, we resampled its UID and observed whether the network’s prediction changed. This process was repeated 10,000 times per node, and the average number of prediction changes was recorded, resulting in an invariance ratio for each node. Finally, we calculated the average invariance ratio across the entire training and test sets to quantify the network’s overall invariance to UIDs. Each evaluation was conducted using five different random seeds.
We evaluate the invariance both on the train set and on the test set because, as we proved in Theorem~\ref{thm:invarinace_can_be_bad}, the network can be invariant to the UIDs for the graphs in the train set but non-invariant to the UIDs for the graphs in the test set.

\begin{table}[h!]
\centering
\small
\caption{Accuracy (\%)$\uparrow$ of \ourmethod~on the isInTriangle task, in interpolation and extrapolation settings. In both settings, SIRI outperforms RNI. }
\label{tab:int_vs_exp}
\begin{tabular}{lcc}
\\
\toprule
  Method $\downarrow$ / Task $\rightarrow$     & Interp. & Extrap. \\ \toprule
  Constant &  75.35 ± 2.09 & 53.70 ± 1.67 \\
RNI    &      74.87   $\pm$ 3.06       &      57.02     $\pm$ 3.39     \\ \midrule
\ourmethod~(Ours)   &      \textbf{88.45 $\pm$ 2.04}         &            \textbf{78.20  $\pm$ 2.53} \\ 

\bottomrule
\end{tabular}
\end{table}

\begin{figure*}[t]
    \centering

      \subfigure[EXP]{\label{figure:exp_1_test_acc} \includegraphics[width=0.45\textwidth]{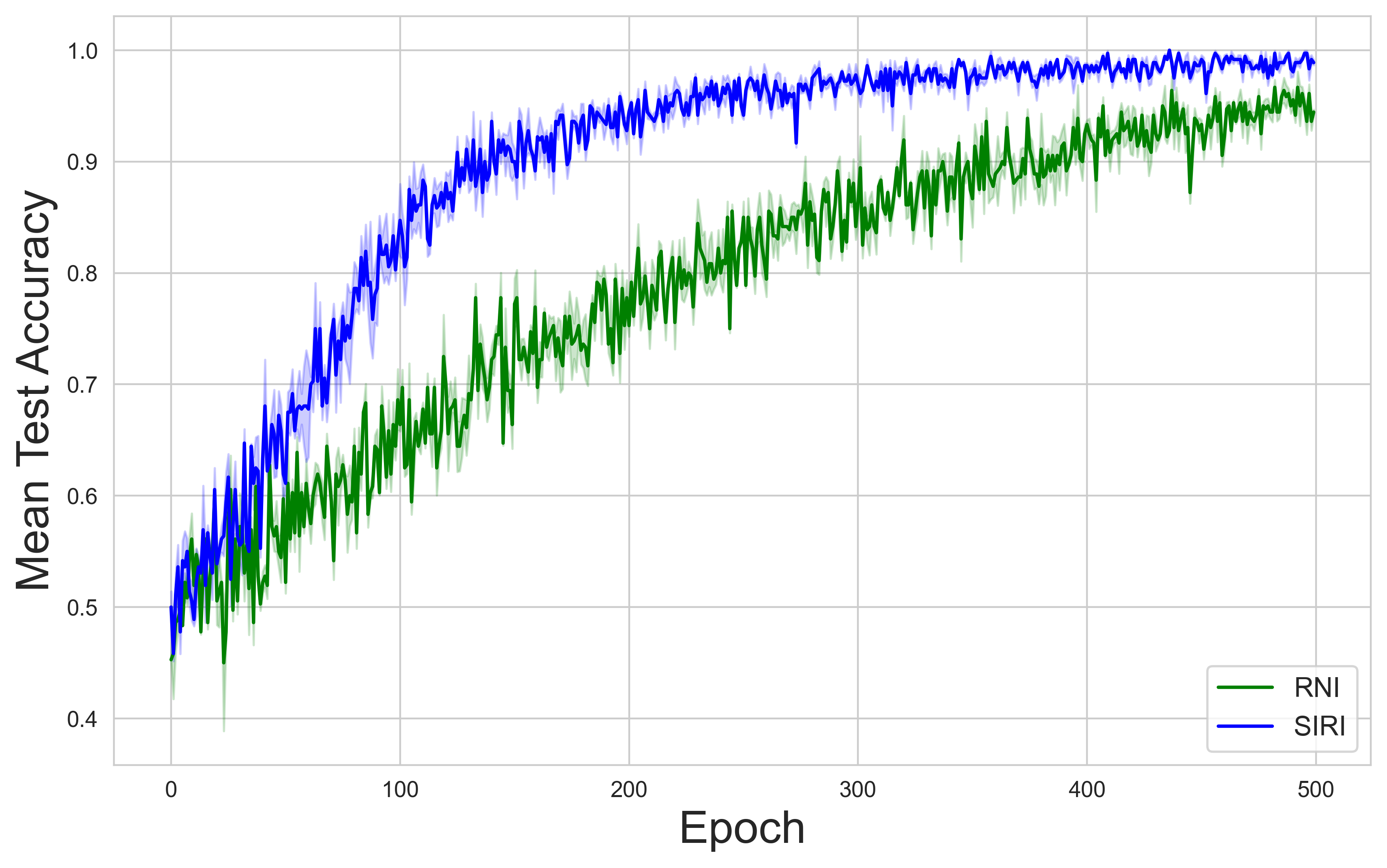}}\quad
      \subfigure[CEXP]{\label{figure:cexp_05_test_acc}\includegraphics[width=0.45\textwidth]{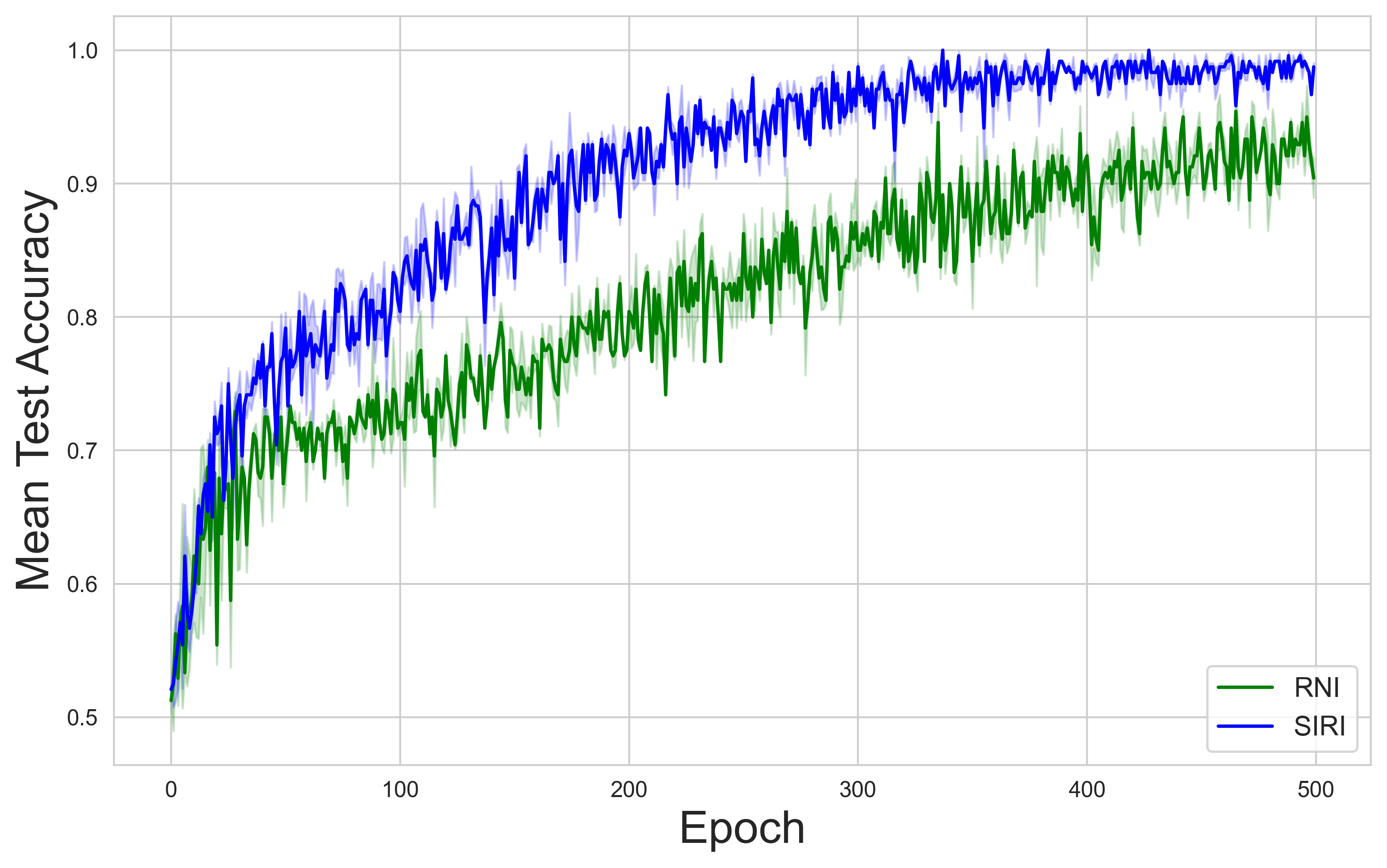}}

    \caption{The learning curves of RNI \ourmethod~on the EXP and CEXP daatasets. While both methods reach almost perfect accuracy, our \ourmethod~offers faster convergence.}
    \label{Figure:exp_cexp}
\end{figure*}

\paragraph{Results}
The interpolation and extrapolation performance of SIRI, compared to RNI, is presented in Table~\ref{tab:int_vs_exp}. 
The network trained with RNI achieves nearly the same accuracy as the baseline (i.e., using a constant non-unique identifier), which is unable to solve the task. This similarity suggests that the RNI-trained network may either be overfitting the UIDs or not leveraging them effectively.
In both settings, SIRI significantly outperforms RNI and the baseline, indicating its effective utilization of UIDs.

Figure~\ref{Figure:siri_vs_rni} illustrates the curves of the invariance ratio during training. For RNI, The network is almost fully UIDs-invariant both the training and test sets.
This persistent high invariance ratio, along with the unchanged accuracy compared to the constant feature baseline, suggests that the RNI-based network did not overfit the UIDs but instead failed to utilize them effectively.
SIRI is also shown to converge to a fully UIDs-invariant solution. Nonetheless, in contrast to RNI, it converged to a UIDs-invariant solution that simultaneously leverages the UIDs and improves generalization, as reflected in our results in \Cref{tab:int_vs_exp}. 
These findings show the ability of SIRI to use RNFs in a way that retains their expressive power while preventing overfitting.


\paragraph{Random UIDs Optimization}
In addition to the contribution of \ourmethod~in enhancing the utilization of random UIDs, we evaluate \ourmethod~coupled with an additional mechanism that optimizes the pairs $R_1, R_2$ to be trained at each iteration, on which we elaborate in \Cref{app:opt_rnfs}. Our results show that by following this strategy combined with \ourmethod, it is possible to obtain further performance gains. Particularly, we obtain 90.10$\pm$1.83 \%  and  79.69$\pm$2.15 \% accuracy on the interpolation and extrapolation settings, respectively. These results indicate that additional approaches to harness the expressive power of UIDs within a network that is also UID invariant can be studied in future works. Importantly, the results presented here show that it is not sufficient to supplement MPGNNs with theoretically expressive node encodings through the form of UIDs, but it is also important to proactively promote the network towards utilizing these encodings, as in our \ourmethod.

\subsection{Improved Convergence Time}

In \citet{abboud2021surprisingpowergraphneural}, the authors showed on the EXP and CEXP datasets that using RNI requires a longer training time compared with a baseline MPGNN. This is because the network relies on implicitly learning UID invariance. We follow their experimental setting to determine whether our \ourmethod~improves  convergence time. 


\paragraph{Datasets} EXP and CEXP \citep{abboud2021surprisingpowergraphneural} contain 600 pairs of graphs (1,200 graphs in total) that cannot be distinguished by 1\&2-WL tests. The goal is to classify each graph into one of two isomorphism classes. Splitting is done by stratified five-fold cross-validation. CEXP is a modified version of EXP, where 50\% of pairs are slightly modified to be distinguishable by 1-WL. It was shown in \cite{abboud2021surprisingpowergraphneural} that a GraphConv \citep{abboud2021surprisingpowergraphneural} with RNI reaches perfect accuracy on the test set.

\paragraph{Setup}
We repeat the protocol from \citet{abboud2021surprisingpowergraphneural}, who evaluated RNI. We use a GraphConv GNN~\citep{morris2021weisfeiler} with \ourmethod~with $8$ layers, $64$ random features, $64$ hidden dimensions, over 500 epochs.

\paragraph{Results}
The learning curves of \ourmethod~and RNI are presented in Figure~\ref{Figure:exp_cexp}. Both methods reach an accuracy of almost 100\%, but \ourmethod~convergence is faster in both tasks, highlighting the improved utilization of UIDs.
\newline\newline 
\begin{table*}[t]

\caption{Pair distinguishing Accuracy (\%)$\uparrow$ on BREC. Overall, SIRI achieves the best total accuracy. With respect to each graph group, SIRI outperforms other random approaches in 2 out of 4 graphs. The largest improvement is achieved on the Regular graphs group. }
\label{tab:brec}
\begin{center}
\begin{tabular}{lcccccccccc}
\\
\toprule
 \multirow{1}{*}{Model $\downarrow$ / Task $\rightarrow$} & \multicolumn{2}{c}{Basic (60)} & \multicolumn{2}{c}{Regular  (140)} & \multicolumn{2}{c}{Extension  (100)} & \multicolumn{2}{c}{CFI  (100)} & \multicolumn{2}{c}{Total (400)}\\
\cmidrule(r{0.5em}){2-3} \cmidrule(l{0.5em}){4-5}\cmidrule(l{0.5em}){6-7}\cmidrule(l{0.5em}){8-9}\cmidrule(l{0.5em}){10-11}
  &  \# & Acc. & \# & Acc.  & \# & Acc.  & \# & Acc.  & \# & Acc.  \\
\midrule
$\,$ DropGNN & 52 & 86.7\% & 41 & 29.3\% & 82 & 82\% & 2 & 2\% & 177 & 44.2\% \\
$\,$ OSAN &  \textbf{56} & \textbf{93.3\% }& 8 & 5.7\% & 79 & 79\% & \textbf{5 }& \textbf{5\%} & 148 & 37\% \\
$\,$ RNI &28 & 46.7\% & 50 & 35.7\% & 83 & 83\%& 0& 0\%& 161&40.3\%\\
\midrule
$\,$ SIRI (Ours) & 40 & 66.7\% & \textbf{100} & \textbf{71.4\%} & \textbf{85}& \textbf{85\%} &1& 1\%& \textbf{226}&\textbf{56.5\%}\\

\bottomrule
\\
\\
\end{tabular}
\end{center}
\vspace{-0.2in}
\end{table*}

\begin{table*}
\caption{Running times and memory costs of the models evaluated on BREC. Our \ourmethod~requires significantly less computation time than other random models, while maintaining a small parameter size.}
\label{tab:cost}
\begin{center}
\begin{tabular}{lcccc}
\toprule
Model & Preprocess Time(s) & Evaluation Time(s) & Total Time(s) & Parameter Size (KB)\\
\midrule
DropGNN & 393 & 533 & 926 & 33 \\
OSAN & $<$1 & 187023 & 187023 & 749\\
RNI & 0& 401& 401& 29\\
\midrule
SIRI (Ours) & 0& 439& 439& 29\\
\bottomrule
\end{tabular}

\end{center}
\end{table*}

\subsection{Improved Expressiveness}

The BREC dataset~\citep{wang2024empiricalstudyrealizedgnn} is a recent dataset introduced to address limitations in evaluating the expressiveness of GNNs. As mentioned in \citet{wang2024empiricalstudyrealizedgnn}, previous works have shown that the commonly used real-life graph benchmarks do not consistently require and benefit from using high-expressiveness models~\citep{pmlr-v97-wu19e, errica2022fair, yang2023graph, bechlerspeicher2024intelligibleeffectivegraphneural}. BREC aims to provide a more comprehensive and challenging benchmark for measuring the ability of GNNs to distinguish between non-isomorphic graphs.

\paragraph{Dataset} The recently proposed \textit{BREC} \citep{wang2024empiricalstudyrealizedgnn} was designed for GNN expressiveness comparison. It addresses the limitations of previous datasets, including difficulty, granularity, and scale, by incorporating 400 pairs of various graphs in four categories (Basic, Regular, Extension, CFI). 
BREC includes 800 non-isomorphic graphs arranged in
a pairwise manner to construct 400 pairs.
BREC comes with an evaluation framework that aims to measure models’ practical ”separation power” directly through a pairwise evaluation method.
The graphs are divided into four groups: 
\newline
\newline
\textbf{Basic Graphs} contain 60 pairs of graphs that are indistinguishable using the 1-WL algorithm. These graphs were identified through an exhaustive search and are specifically crafted to be non-regular. They can be viewed as an expansion of the EXP dataset, offering a similar level of difficulty. However, this dataset provides a larger number of instances and features more complex graph structures.
\newline
\newline
\textbf{Regular graphs} include 140 pairs of regular graphs, categorized into simple regular graphs, strongly regular graphs, 4-vertex condition graphs, and distance regular graphs. In these regular graphs, every node has the same degree. They are indistinguishable by the 1-WL algorithm, and some studies have explored GNN expressiveness from this perspective~\citep{li2020distanceencodingdesignprovably, zhang2021nestedgraphneuralnetworks}.
\newline
\newline
\textbf{Extension graphs} consist of 100 pairs of graphs inspired by \citet{papp2022theoreticalcomparisongraphneural}. Unlike previous parts, these extensions do not have a strict comparative relationship with one another. They include graphs that range in difficulty from being indistinguishable by 1-WL up to 3-WL, thereby improving the granularity of distinguishing difficulty levels
\newline
\newline
\textbf{CFI graphs}  includes 100 pairs of graphs inspired by \citet{63543}, with difficulty levels reaching up to 4-WL indistinguishability. This most challenging section pushes the upper limits of difficulty even further. Additionally, the graphs in this part are larger than those in other sections, containing up to 198 nodes.

\paragraph{Setup}
The evaluation protocol of BREC, as proposed in \citet{wang2024empiricalstudyrealizedgnn}, adheres to the Siamese network design to train a model to distinguish each pair of graphs. For a pair of graphs inputted, it outputs a pair of embeddings. Subsequently, the difference between them is assessed using cosine similarity.
We evaluate BREC on the GraphConv model with RNI and a GraphConv model with \ourmethod. We compare it to other randomization GNNs that were previously evaluated in \citet{wang2024empiricalstudyrealizedgnn}. We followed the hyper-parameters search of DropGNN, as done in \citet{wang2024empiricalstudyrealizedgnn}.\newline
We also provide a running time and model size following \citet{wang2024empiricalstudyrealizedgnn}.
More details on the training protocol that follows  \citet{wang2024empiricalstudyrealizedgnn}, and the chosen hyperparameters are provided in the Appendix.

\paragraph{Results}
The performances obtained on the BREC dataset tasks are summarized in Table~\ref{tab:brec}. Each column corresponds to one of the four graph groups, and the last column shows the overall performance over the whole dataset. In each column, we present the number of graphs in the group that were separated, and its corresponding percentage over the group size, which also appears in parentheses in each column title. \newline
Overall, our \ourmethod~demonstrates improved performance over RNI in all graph groups. This demonstrates its contribution to improving the  GNN's ability to harness the enhanced expressiveness of UIDs. \newline
\ourmethod~improved the total accuracy in 12.3\%.
With respect to each group of graphs, \ourmethod~outperformed all other evaluated methods in  2 out of 4 graph groups, where the largest improvement is achieved in the Regular graphs group.

We also underscore the efficient running times of \ourmethod~as presented in Table~\ref{tab:cost}.
SIRI significantly boosts performance while incurring only a slight cost increase over RNI. Besides the low runtimes required by \ourmethod, it also maintains a small model size compared with other methods.
Importantly, although the OSAN method outperforms SIRI in two of the four graph groups, it is orders of magnitude more costly.


\section{Future Work}
 We believe this work opens up many new avenues for exploration. In this section, we discuss several future research directions.

An intriguing open question is whether two layers of a GNN with unique identifiers (UIDs) suffice to achieve additional expressiveness beyond the 1-1-WL test. In Section~\ref{sec:theory} we demonstrated that three layers already provide greater expressiveness than 1-WL.\newline Determining whether this can be accomplished with only two layers remains unresolved.

Furthermore, we showed that using only UIDs-invariant layers does not enhance expressiveness, whereas enforcing invariance solely in the final layer does. 
\newline
Fully understanding the necessary limitations on layers regarding UIDs-invariance, and identifying which functions can be computed under various settings, presents an interesting direction for future research.

Our experiments show that the utilization of UIDs can be improved with \ourmethod, while \textit{learning} to produce an invariant network. Another interesting open question is how to design GNNs that benefit from the added theoretical expressiveness of UIDs \textit{and} are invariant to UIDs \textit{by design}.

We conjecture that this can be achieved by combining a matching oracle with a GNN architecture. 
A matching oracle is a function that takes two nodes from the graph as input and returns 1 if they are the same node and 0 otherwise. \newline
The following Theorem shows that any UIDs-invariant function can be computed with such an oracle. 
\begin{theorem}[informal]\label{thm:matching_oracle}
      Any UIDs-invariant function can be computed with a matching oracle.
\end{theorem}

\begin{conjecture}[informal]
  There exists a GNN that uses a matching oracle with UIDs and can compute functions that are not 1-WL.
\end{conjecture}
Nonetheless, it is an open question how to combine such an oracle with, e.g., Message-Passing GNNs in an effective way that also results in an improved empirical performance.

\section{Conclusions}
In this paper, we harness the theoretical power 
of unique node identifiers (UIDs) and improve the ability of Graph Neural Networks (GNNs) to utilize them effectively.
We provide a theoretical analysis that lays a strong foundation for designing an approach that explicitly enforces invariance to UIDs in GNNs while taking advantage of their expressive power to offer improved downstream performance. \newline Building upon our theoretical results, we propose an explicit regularization technique through a contrastive loss, an approach that we call \ourmethod.
 Through comprehensive experiments, we demonstrate that \ourmethod~not only significantly enhances generalization and extrapolation by enabling the network to utilize UIDs effectively but also accelerates convergence compared to existing methods. These results highlight \ourmethod~as an efficient solution for boosting the expressiveness of GNNs through UIDs utilization, making it a valuable tool for use with any GNN.

\section*{Acknowledgements}
This work was supported by the Tel Aviv University Center for AI and Data Science (TAD) and the Israeli Science Foundation grants 1186/18 and 1437/22.
\bibliography{biblio}

\begin{thebibliography}{}

\bibitem[Abboud et~al., 2021]{abboud2021surprisingpowergraphneural}
Abboud, R., İsmail~İlkan Ceylan, Grohe, M., and Lukasiewicz, T. (2021).
\newblock The surprising power of graph neural networks with random node initialization.

\bibitem[Barabasi and Albert, 1999]{badist}
Barabasi, A.-L. and Albert, R. (1999).
\newblock Emergence of scaling in random networks.
\newblock {\em Science}, 286(5439):509--512.

\bibitem[Bechler-Speicher et~al., 2024]{bechlerspeicher2024intelligibleeffectivegraphneural}
Bechler-Speicher, M., Globerson, A., and Gilad-Bachrach, R. (2024).
\newblock The intelligible and effective graph neural additive networks.

\bibitem[Benton et~al., 2020]{benton2020learning}
Benton, G., Finzi, M., Izmailov, P., and Wilson, A.~G. (2020).
\newblock Learning invariances in neural networks.
\newblock In {\em Proceedings of NeurIPS}.

\bibitem[Bevilacqua et~al., 2022]{bevilacqua2022equivariant}
Bevilacqua, B., Frasca, F., Lim, D., Srinivasan, B., Cai, C., Balamurugan, G., Bronstein, M.~M., and Maron, H. (2022).
\newblock Equivariant subgraph aggregation networks.

\bibitem[Cai et~al., 1989]{63543}
Cai, J.-Y., Furer, M., and Immerman, N. (1989).
\newblock An optimal lower bound on the number of variables for graph identification.
\newblock In {\em 30th Annual Symposium on Foundations of Computer Science}, pages 612--617.

\bibitem[Cant\"{u}rk et~al., 2024]{gpse_canturk}
Cant\"{u}rk, S., Liu, R., Lapointe-Gagn\'{e}, O., L\'{e}tourneau, V., Wolf, G., Beaini, D., and Ramp\'{a}\v{s}ek, L. (2024).
\newblock Graph positional and structural encoder.
\newblock In Salakhutdinov, R., Kolter, Z., Heller, K., Weller, A., Oliver, N., Scarlett, J., and Berkenkamp, F., editors, {\em Proceedings of the 41st International Conference on Machine Learning}, volume 235 of {\em Proceedings of Machine Learning Research}, pages 5533--5566. PMLR.

\bibitem[Chen et~al., 2023]{chen2023equivalencegraphisomorphismtesting}
Chen, Z., Villar, S., Chen, L., and Bruna, J. (2023).
\newblock On the equivalence between graph isomorphism testing and function approximation with gnns.

\bibitem[Eliasof et~al., 2024]{eliasof2024granola}
Eliasof, M., Bevilacqua, B., Sch{\"o}nlieb, C.-B., and Maron, H. (2024).
\newblock Granola: Adaptive normalization for graph neural networks.
\newblock In {\em Proceedings of NeurIPS}.

\bibitem[Eliasof et~al., 2023]{eliasof2023graph}
Eliasof, M., Frasca, F., Bevilacqua, B., Treister, E., Chechik, G., and Maron, H. (2023).
\newblock Graph positional encoding via random feature propagation.
\newblock In {\em International Conference on Machine Learning}, pages 9202--9223. PMLR.

\bibitem[Errica et~al., 2022]{errica2022fair}
Errica, F., Podda, M., Bacciu, D., and Micheli, A. (2022).
\newblock A fair comparison of graph neural networks for graph classification.

\bibitem[Garg et~al., 2020a]{garg2020generalizationrepresentationallimitsgraph}
Garg, V.~K., Jegelka, S., and Jaakkola, T. (2020a).
\newblock Generalization and representational limits of graph neural networks.

\bibitem[Garg et~al., 2020b]{cyclesgnns}
Garg, V.~K., Jegelka, S., and Jaakkola, T. (2020b).
\newblock Generalization and representational limits of graph neural networks.

\bibitem[Jaderberg et~al., 2015]{jaderberg2015spatial}
Jaderberg, M., Simonyan, K., Zisserman, A., and Kavukcuoglu, K. (2015).
\newblock Spatial transformer networks.
\newblock In {\em Advances in neural information processing systems}, pages 2017--2025.

\bibitem[Jia et~al., 2024]{jia2024graph}
Jia, T., Li, H., Yang, C., Tao, T., and Shi, C. (2024).
\newblock Graph invariant learning with subgraph co-mixup for out-of-distribution generalization.
\newblock In {\em Proceedings of the AAAI Conference on Artificial Intelligence}, volume~38, pages 8562--8570.

\bibitem[Laptev et~al., 2016]{laptev2016tipooling}
Laptev, D., Savinov, N., Buhmann, J.~M., and Pollefeys, M. (2016).
\newblock Ti-pooling: transformation-invariant pooling for feature learning in convolutional neural networks.
\newblock In {\em Proceedings of the IEEE Conference on Computer Vision and Pattern Recognition}, pages 289--297.

\bibitem[Li et~al., 2020]{li2020distanceencodingdesignprovably}
Li, P., Wang, Y., Wang, H., and Leskovec, J. (2020).
\newblock Distance encoding: Design provably more powerful neural networks for graph representation learning.

\bibitem[Loukas, 2020]{loukas2020graphneuralnetworkslearn}
Loukas, A. (2020).
\newblock What graph neural networks cannot learn: depth vs width.

\bibitem[Morris et~al., 2023]{morris2023weisfeiler}
Morris, C., Lipman, Y., Maron, H., Rieck, B., Kriege, N.~M., Grohe, M., Fey, M., and Borgwardt, K. (2023).
\newblock Weisfeiler and leman go machine learning: The story so far.
\newblock {\em The Journal of Machine Learning Research}, 24(1):15865--15923.

\bibitem[Morris et~al., 2021]{morris2021weisfeiler}
Morris, C., Ritzert, M., Fey, M., Hamilton, W.~L., Lenssen, J.~E., Rattan, G., and Grohe, M. (2021).
\newblock Weisfeiler and leman go neural: Higher-order graph neural networks.

\bibitem[Murphy et~al., 2019]{murphy2019relational}
Murphy, R.~L., Rao, V., Rincon, B., and Ribeiro, B. (2019).
\newblock Relational pooling for graph representations.
\newblock In {\em International Conference on Machine Learning}, pages 4663--4673. PMLR.

\bibitem[Papp et~al., 2021]{papp2021dropgnnrandomdropoutsincrease}
Papp, P.~A., Martinkus, K., Faber, L., and Wattenhofer, R. (2021).
\newblock Dropgnn: Random dropouts increase the expressiveness of graph neural networks.

\bibitem[Papp and Wattenhofer, 2022]{papp2022theoreticalcomparisongraphneural}
Papp, P.~A. and Wattenhofer, R. (2022).
\newblock A theoretical comparison of graph neural network extensions.

\bibitem[Qian et~al., 2022]{qian2022ordered}
Qian, C., Rattan, G., Geerts, F., Niepert, M., and Morris, C. (2022).
\newblock Ordered subgraph aggregation networks.
\newblock In Oh, A.~H., Agarwal, A., Belgrave, D., and Cho, K., editors, {\em Advances in Neural Information Processing Systems}.

\bibitem[Sato et~al., 2021]{sato2021randomfeaturesstrengthengraph}
Sato, R., Yamada, M., and Kashima, H. (2021).
\newblock Random features strengthen graph neural networks.

\bibitem[Wang and Zhang, 2024a]{wang2024empiricalstudyrealizedgnn}
Wang, Y. and Zhang, M. (2024a).
\newblock An empirical study of realized gnn expressiveness.

\bibitem[Wang and Zhang, 2024b]{wang2024an}
Wang, Y. and Zhang, M. (2024b).
\newblock An empirical study of realized {GNN} expressiveness.
\newblock In {\em Forty-first International Conference on Machine Learning}.

\bibitem[Wu et~al., 2019]{pmlr-v97-wu19e}
Wu, F., Souza, A., Zhang, T., Fifty, C., Yu, T., and Weinberger, K. (2019).
\newblock Simplifying graph convolutional networks.
\newblock In Chaudhuri, K. and Salakhutdinov, R., editors, {\em Proceedings of the 36th International Conference on Machine Learning}, volume~97 of {\em Proceedings of Machine Learning Research}, pages 6861--6871. PMLR.

\bibitem[Xia et~al., 2023]{xia2023learning}
Xia, D., Wang, X., Liu, N., and Shi, C. (2023).
\newblock Learning invariant representations of graph neural networks via cluster generalization.
\newblock In {\em Thirty-seventh Conference on Neural Information Processing Systems}.

\bibitem[Xu et~al., 2019]{gin}
Xu, K., Hu, W., Leskovec, J., and Jegelka, S. (2019).
\newblock How powerful are graph neural networks?
\newblock In {\em International Conference on Learning Representations}.

\bibitem[Yang et~al., 2023]{yang2023graph}
Yang, C., Wu, Q., Wang, J., and Yan, J. (2023).
\newblock Graph neural networks are inherently good generalizers: Insights by bridging gnns and mlps.

\bibitem[You et~al., 2021]{you2021identity}
You, J., Ying, R., and Leskovec, J. (2021).
\newblock Identity-aware graph neural networks.
\newblock In {\em Proceedings of the AAAI Conference on Artificial Intelligence}, volume~35, pages 10737--10745.

\bibitem[Zhang and Li, 2021]{zhang2021nestedgraphneuralnetworks}
Zhang, M. and Li, P. (2021).
\newblock Nested graph neural networks.

\end{thebibliography}

 \clearpage

 \appendix

\onecolumn
\section{Proofs}
\subsection{Proof of Theorem~\ref{thm:invarinace_can_be_bad}}
We will show that there exists a function $f$  that is invariant to UIDs with respect to one set of graphs \( S \) and non-invariant with respect to another set of graphs \( S' \). 

For a given graph $g=(V,E)$, we denote the unique identifier of node $v$ as \( \text{UID}_v \) for each node \( v \in V \).
Now consider the following function:

   \[
   f(g) = 
   \begin{cases} 
   P(g) & \text{if } g \in S \\ 
   \sum_{v \in V} \text{UID}_v & \text{else }
   \end{cases}
   \]

$f$ computes a function $P$ over the graphs in $S$, and returns the sum of UIDs of $g$ otherwise. 
Here, $P$ is any graph property that is UIDs-invariant, such as the existence of the Eulerian path.
Summing the UIDs is not invariant to the UIDs' values. Therefore, $f$ is invariant to UIDs with respect to graphs in $S$ but not with respect to the graphs in $S$.

\subsection{Proof of direct implication of Theorem~\ref{thm:invarinace_can_be_bad}}
In the main text we mention that a direct implication of Theorem~\ref{thm:invarinace_can_be_bad} is that a GNN can be UIDs-invariant to a train set and non UIDs-invariant to a test set. 
This setting only differs from the setting in Theorem~\ref{thm:invarinace_can_be_bad} in that now the learned function is restricted to one that can be realized by a GNN.
Indeed, as proven in~\citet{abboud2021surprisingpowergraphneural}, a GNN with UIDs is a universal approximator. Specifically, with enough layers and width, a MPGNN can reconstruct the adjacency matrix in some hidden layer, and then compute over it the same function as in the proof of Theorem~\ref{thm:invarinace_can_be_bad}.

\subsection{Proof of Theorem~\ref{thm:mpnnwithidsepxressivity}}
To prove the Theorem, we will show that a GNN-R that is UID invariant in every layer is equivalent to a regular GNN with identical constant values on the nodes. We focus on the case where no node features exist in the data. For the sake of simplicity, we assume, without loss of generality, that the UIDS are of dimension $1$ and that the fixed constant features are of the value $1$. We focus on Message-Passing GNNs, but for the sake of conciseness, we will not repeat that in the proof.
Let $G$ be a GNN-R with $L$ layers that is UIDs-invariant in every layer. Let $G'$ be a regular GNN with $L$ layers that assigns constant $1$ features to all nodes. 
We denote a model up to its $l$'th layer as $A^{(l)}$.
We will prove by induction on the layers $l\in \{0,...,L-1\}$ that the output of every layer $H^{(l)}, l>0$ can be realized by the corresponding layers in $G'$.
we denote the inputs after the assignments of values to nodes by the networks as $h'^{0}$ for $G$ and  $h'^{0}$ for $G'$.

Base Case - $l=0$: we need to show that there is $G'^{(1)}$ such that $H'^{(1)} =  H^{(1)}$
As $l=0$ is a  UIDs-invariant layer, 
$G^{(1)}(h^{0}) = G^{(1)}(\bar{h}^{0}$ for any $\bar{h} \neq h$.
Specifically, $G^{(1)}(h^{0}) = G^{(1)}(h'^{0})$
Therefore we can have $G^{(1)} = G'^{(1)}$ and then $H'^{(1)} =  H^{(1)}$.

Assume the statement is true up to the $L-2$ layer.
We now prove that there is $G^{(L-1)}$ such that $H^{(L-1)} = H'^{(L-1)}$.

From the inductive assumption, there is $G'{(L-2)}$ such that $H^{(L-2)} = H'^{(L-2)}$. Let  us take such $G'{(L-2)}$. As the $L$'th layer is UIds-invariant, it holds that $G^{(L-2)}(h^{(L-1})) = G^{(L-2)}(H'^{(L-2)})$. Therefore, we can have $G'^{(L-2)} = G^{(L-2)}$ and then $H^{(L-1)} = H'^{(L-1)}$.

\subsection{Proof of Theorem~\ref{thm:only_last_layer}}
To prove Theorem~\ref{thm:only_last_layer}, we will construct a GNN with three layers, where the first two layers are non UIDs-invariant, and the last layer is UIDs-invariant, that solves the isInTriangle task. It was already shown that isInTriangle cannot be solved by 1-WL GNNs~\citep{cyclesgnns}.

Let $G$ be a graph with $n$ nodes, each assigned a unique UID of dimension $d$, and assume no other node features are associated with the nodes of $G$. 

In every layer $i$, $f^{(i)}$, each node $v$ updates its representation $h_v^{(i-1)}$ by computing a message using a function $m^{(i)}$, aggregates the messages from its neighbors $N(v)$, $\{m^{(i)}(u)\}_{u \in N(v)}$, using a function $agg^{(i)}$, and updates its representation $h_v^{(i)} = \text{update}^{(i)}$ by combining the outputs of $m^{(i)}$ and $agg^{(i)}$.

We define $f$ as follows. The inputs are $h_v^{(0)} = \text{UID}_v$. In the first two layers, the message function simply writes the UID of the node by copying it from the current representation, $m^{(i)} = h_v^{(0)}[0]$. Then it aggregates the messages from neighbors by concatenating them in arbitrary order: $agg^{(i)} = \text{CONCAT}(\{m^{(i)}(u)\}_{u \in N(v)})$. Then the update function concatenates the node's message and its aggregation output: $\text{update}^{(i)} = \text{CONCAT}(m^{(i)}, agg^{(i)})$. Therefore, in the output of the second layer, the node's own UID is in position $0$ of the representation vector. Notice that the first and second layers are not UIDs-invariant, as replacing the UIDs will result in a different vector.

In the final layer, the message and aggregate functions perform the same operations as in the previous layers, but the update layer matches the output of the message function with the entries in the output of the aggregation function to check if the UID of the node appears in the messages from the neighbors. If it does, it outputs 1; otherwise, it outputs 0. This layer is UIDs-invariant, as its outputs depend on the re-appearance of the same UIDs, without dependency on their values.

\subsection{Proof of Theorem~\ref{thm:isomorphism}}
We will show that unless Graph Isomorphism (GI) is not NP-Complete, a GNN-R that is UIDs-invariant runs and runs in polynomial time, cannot be a universal approximator. 
It was previously shown in \cite{chen2023equivalencegraphisomorphismtesting} that a universal approximation of equivariant graph functions, can solve GI. Therefore, if GNN-R is a UIDs-invariant and runs in polynomial time, GI can be solved in polynomial time, and therefore, GI is in $P$. Therefore, it cannot be NP-complete, unless $P=NP$. 

\subsection{Proof of Theorem~\ref{thm:matching_oracle}}
Let \( f \) be an equivariant function of graphs that is also UIDs-invariant. We will demonstrate that \( f \) can be expressed using a matching oracle. Let \( o \) be a matching oracle defined as follows:

\[
o(u, v) = 
\begin{cases} 
1 & \text{if } u = v \\ 
0 & \text{otherwise}
\end{cases}
\]

Assume we have a serial program that computes \( f \). We will compute \( f \) using a serial function \( g \) that utilizes the oracle \( o \). The function \( g \) incorporates caching  Let \( \text{Cache} \) denote a cache that stores nodes with an associated value to each node.

The function \( g \) operates as $f$, except for the follows:

\begin{enumerate}[label=(\alph*)]
    \item When \( f \) needs to access the UID of a node \( x \), $g$ checks whether \( x \) already exists in the cache by matching \( x \) with each node stored in the cache using the oracle \( o \).
    
    \item If \( x \) is found in the cache, \( g \) retrieves and returns the value associated with it.
    
    \item If \( x \) is not found in the cache, \( g \) adds it to the cache and assigns a new value to \( x \) as \( \text{UID}(x) = \text{size}(\text{Cache}) + 1 \), and return its value.
\end{enumerate}

By the assumption that \( f \) is invariant under UIDs, we have \( g = f \).

 \section{Experimental Details}
 \label{app:experimental_settings}

 \paragraph{isInTriangle}
 We trained a 6-layer GraphConv for $2000$ epochs with Adam optimizer, $64$ random features, and $64$ hidden layers with ReLU activations.

  \paragraph{EXP and CEXP}
  We followed the protocol of \citet{abboud2021surprisingpowergraphneural} and used Adam optimizer with a learning rate of $1e-4$. The network has $8$ GraphConv layers followed by a sum-pooling and a $3$ layer readout function. All hidden layers have $64$ dimensions, and we used $64$ random features and we did not discard the original features in the data. 

   \paragraph{BREC}
   For the BREC dataset, we conducted a search over other hyperparameters following what was done in~\citet{wang2024an}. 
   We used the Adam optimizer with a learning rate searched from $\{1e-3, 1e-4, 1e-5\}$, weight decay selected from
$\{1e-3, 1e-4, 1e-5\}$, and batch size chosen from $\{8, 16, 32\}$. We used $100$ epochs as in the evaluated DropGNN model from ~\citet{wang2024an},
 We trained an 8-layer GraphConv with SIRI and $64$ random features and $64$ hidden layers with ReLU activations.

\subsection{Furthur RNF optimization}\label{app:opt_rnfs}

Here, we present an evaluation of the isInTriangle task, with additional RNF optimization as described at the end of Section~\ref{sec:method}.

Instead of randomly selecting $R_2$ for the contrastive loss, we sample 5 vectors of $R_2$ and apply the gradient
update with the one that maximizes the contrastive loss.
The results are presented in Table~\ref{tab:int_vs_exp_with_opt}. The additional optimization does improve both interpolation and extrapolation with respect to randomly selecting one $R_2$.

\begin{table}[h!]
\centering
\small
\caption{Accuracy (\%)$\uparrow$ of \ourmethod~on the isInTriangle task, in interpolation and extrapolation settings. In both settings, SIRI outperforms RNI. }
\label{tab:int_vs_exp_with_opt}
\begin{tabular}{lcc}
\\
\toprule
  Method $\downarrow$ / Task $\rightarrow$     & Interp. & Extrap. \\ \toprule
  Constant &  75.35 ± 2.09 & 53.70 ± 1.67 \\
RNI    &      74.87   $\pm$ 3.06       &      57.02     $\pm$ 3.39     \\ \midrule
\ourmethod~(Ours)   &      \textbf{88.45 $\pm$ 2.04}         &            \textbf{78.20  $\pm$ 2.53} \\ 
\ourmethod~+ OPT (Ours) &\textbf{90.10 ± 1.83}      &    \textbf{79.69  $\pm$ 2.15} \\ 
\bottomrule
\end{tabular}
\end{table}

\subsection{Additional results for BREC}
 In ~\citet{wang2024an}, the authors evaluated expressiveness over a wide range of methods from different types; some are not GNN-based. In the main paper, we compare our method to random-based GNNs, yet here we wish to provide a wider comparison. This is presented in Table~\ref{tab:experiment_brec_all}.
 It is important to note that these additional methods we include here, are highly costly in running time and memory, as shown in Table~\ref{tab:full_cost}

\begin{table*}
\caption{Pair distinguishing accuracies on BREC}
\label{tab:experiment_brec_all}
\begin{center}
\resizebox{1.0\textwidth}{!}{
\begin{tabular}{lccccccccccc}
\toprule
~ & ~ & \multicolumn{2}{c}{Basic Graphs (60)} & \multicolumn{2}{c}{Regular Graphs (140)} & \multicolumn{2}{c}{Extension Graphs (100)} & \multicolumn{2}{c}{CFI Graphs (100)} & \multicolumn{2}{c}{Total (400)}\\
\cmidrule(r{0.5em}){3-4} \cmidrule(l{0.5em}){5-6}\cmidrule(l{0.5em}){7-8}\cmidrule(l{0.5em}){9-10}\cmidrule(l{0.5em}){11-12}
Type & Model &  Number & Accuracy & Number & Accuracy  & Number & Accuracy  & Number & Accuracy  & Number & Accuracy  \\
\midrule
\multirow{7}{*}{Non-GNNs}
& 3-WL & 60 & 100\%  & 50 & 35.7\% & 100 & 100\%  & 60 & 60.0\%  & 270 & 67.5\% \\
& SPD-WL & 16 & 26.7\% & 14 & 11.7\% & 41 & 41\% & 12 & 12\% & 83 & 20.8\% \\
& $S_3$ & 52 & 86.7\% & 48 & 34.3\% & 5 & 5\% & 0 & 0\% & 105 & 26.2\%    \\
& $S_4$ & 60 & 100\% & 99 & 70.7\%  & 84 & 84\% & 0 & 0\% & 243 & 60.8\% \\
& $N_1$ & 60 & 100\% & 99 & 85\% & 93 & 93\% & 0 & 0\% & 252 & 63\% \\
& $N_2$ & 60 & 100\% & 138 & 98.6\% & 100 & 100\% & 0 & 0\% & 298 & 74.5\% \\
& $M_1$ & 60 & 100\% & 50 & 35.7\%  & 100 & 100\% & 41 & 41\% & 251 & 62.8\% \\
\midrule
\multirow{9}{*}{Subgraph GNNs}
& NGNN & 59 & 98.3\% & 48 & 34.3\% & 59 & 59\%  & 0 & 0\%  & 166 & 41.5\%  \\
& DE+NGNN & 60 & 100\%& 50 & 35.7\% & 100 & 100\%& 21 & 21\% & 231 & 57.8\%\\
& DS-GNN & 58 & 96.7\% & 48 & 34.3\%  & 100 & 100\%  & 16 & 16\%  & 222 & 55.5\% \\
& DSS-GNN & 58 & 96.7\% & 48 & 34.3\% & 100 & 100\% & 15 & 15\%  & 221 & 55.2\% \\
& SUN & 60 & 100\% & 50 & 35.7\% & 100 & 100\% & 13 & 13\% & 223 & 55.8\% \\
& SSWL\_P & 60 & 100\% & 50 & 35.7\% & 100 & 100\% & 38 & 38\% & 248 & 62\% \\
& GNN-AK & 60 & 100\% & 50 & 35.7\% & 97 & 97\%& 15 & 15\% & 222 & 55.5\%\\
& KP-GNN & 60 & 100\% & 106 & 75.7\%& 98 & 98\%& 11 & 11\% & 275 & 68.8\% \\ 
& I$^2$-GNN & 60 & 100\%& 100 & 71.4\% & 100 & 100\%& 21 & 21\% & 281 & 70.2\%\\
\midrule
\multirow{3}{*}{k-WL GNNs}
& PPGN & 60 & 100\% & 50 & 35.7\% & 100 & 100\% & 23 & 23\%  & 233 & 58.2\% \\
& $\delta$-k-LGNN & 60 & 100\% & 50 & 35.7\%& 100 & 100\%& 6 & 6\% & 216 & 54\% \\ 
& KC-SetGNN & 60 & 100\% & 50 & 35.7\%& 100 & 100\%& 1 & 1\% & 211 & 52.8\% \\
\midrule
\multirow{1}{*}{Substructure GNNs}
& GSN & 60 & 100\% & 99 & 70.7\%  & 95 & 95\% & 0 & 0\% & 254 & 63.5\% \\
\midrule
\multirow{1}{*}{Transformer GNNs}
& Graphormer & 16 & 26.7\% & 12 & 8.6\% & 41 & 41\% & 10 & 10\% & 79 & 19.8\% \\
\midrule
\multirow{4}{*}{Random GNNs}
& DropGNN & 52 & 86.7\% & 41 & 29.3\% & 82 & 82\% & 2 & 2\% & 177 & 44.2\% \\
& OSAN &  56 & 93.3\% & 8 & 5.7\% & 79 & 79\% & 5 & 5\% & 148 & 37\% \\
 &RNI &28 & 46.7\% & 50 & 35.7\% & 83 & 83\%& 0& 0\%& 161&40.3\%\\
&SIRI (Ours) & 40 & 66.7\% &100 & 71.4\% &85& 85\% &1& 1\%& 226&56.5\%\\

\bottomrule
\end{tabular}
}
\end{center}
\vspace{-0.2in}
\end{table*}

 \begin{table}
\caption{Model costs}
\vspace{0.2in}
\label{tab:full_cost}
\begin{center}
\begin{tabular}{lcccc}
\toprule
Model & Preprocess Time(s) & Evaluation Time(s) & Total Time(s) & Parameter Size (KB)\\

\midrule
NGNN & 516 & 388 & 904 & 14 \\
DE+NGNN &3123 & 1087 & 4200 & 787\\
DS-GNN & 2343 & 3300 & 5643 & 173\\
DSS-GNN & 2343 & 1008 & 3351 & 162 \\
SUN & 2343 & 1925 & 4268 & 748 \\
SSWL\_P & $>>$644 & $>>$4085 & $>>$4729 & 924 \\
GNN-AK &768 & 1048 & 1816 & 4742\\
KP-GNN &73113 & 873 & 73986 & 366 \\
I$^2$GNN & $>$10460 & $>$7013 & $>$17473 & 236\\
PPGN &47 & 156 & 477 & 90 \\
$\delta$-k-LGNN & 5 & 2532 & 2537 & 62 \\
KC-SetGNN &$>>$118005 & $>>$5670 & $>>$123675 & 1416 \\
GSN &$>$5231 & $>$150 & $>$5381 & 431 \\
Graphormer & $>>$66 & $>>$13681 & $>>$13747 & 3667\\
DropGNN & 393 & 533 & 926 & 33 \\
OSAN & $<$1 & 187023 & 187023 & 749\\
RNI & 0& 401& 401& 29\\
\midrule
SIRI (Ours) & 0& 439& 439& 29\\

\bottomrule
\end{tabular}

\end{center}
\end{table}

\end{document}